\documentclass[11pt,a4paper]{article}
\usepackage[hyperref]{acl2021}
\usepackage{times}
\usepackage{latexsym}

\usepackage[T1]{fontenc}

\usepackage[utf8]{inputenc}

\usepackage{xr-hyper,refcount}
\usepackage{graphicx}
\usepackage[T1]{fontenc}    
\usepackage{url}            
\usepackage{booktabs}       
\usepackage{amsfonts}       
\usepackage{nicefrac}       
\usepackage{microtype}      
\usepackage[parfill]{parskip}
\usepackage{algorithm,algorithmic}
\usepackage{amsmath,amsthm,amssymb,bbm}
\usepackage{mathtools}
\usepackage{cases}
\usepackage{comment}
\usepackage{subcaption}
\usepackage{algorithm,algorithmic}
\usepackage{color}
\usepackage{xcolor}
\usepackage{appendix}
\usepackage{xspace}
\usepackage[shortlabels]{enumitem}
\usepackage[english]{babel}
\usepackage{bm}
\usepackage{arydshln}
\usepackage{makecell}
\usepackage{multirow}
\usepackage{framed}
\usepackage{caption}
\usepackage[scientific-notation=true]{siunitx}
\usepackage{wrapfig}
\usepackage{lipsum}
\usepackage{sidecap}
\usepackage{pifont}
\usepackage{tikz}
\theoremstyle{plain}
\usepackage{fixltx2e}
\usepackage[flushleft]{threeparttable}
\usepackage{diagbox}
\usepackage{scalerel}

\aclfinalcopy

\newcommand{\namel}{\textsc{Noisy Identity MLP}}
\newcommand{\names}{\textsc{NI-MLP}}

\DeclareMathOperator*{\argmin}{arg\,min}

\newcommand\blfootnote[1]{%
  \begingroup
  \renewcommand\thefootnote{}\footnote{#1}%
  \addtocounter{footnote}{-1}%
  \endgroup
}

\definecolor{gg}{RGB}{15,150,15}
\definecolor{rr}{RGB}{230,45,45}

\title{Learning Language and Multimodal Privacy-Preserving Markers\\of Mood from Mobile Data}

\author{%
  Paul Pu Liang$^{1\star}$, Terrance Liu$^{1\star}$, Anna Cai$^1$, Michal Muszynski$^1$, Ryo Ishii$^1$,\\ {\bf Nicholas Allen$^2$, Randy Auerbach$^3$, David Brent$^4$,}\\
  {\bf Ruslan Salakhutdinov$^1$, Louis-Philippe Morency$^1$}\\
  $^1$Carnegie Mellon University \ $^2$University of Oregon\\
  $^3$Columbia University \ $^4$University of Pittsburgh\\
  {\small\texttt{\{pliang,terrancl,annacai,mmuszyns,rishii,rsalakhu,morency\}@cs.cmu.edu}}\\
  {\small\texttt{nallen3@uoregon.edu \ rpa2009@cumc.columbia.edu \ brentda@upmc.edu}}
}

\date{}

\begin{document}
\maketitle

\begin{abstract}
Mental health conditions remain underdiagnosed even in countries with common access to advanced medical care. The ability to accurately and efficiently predict mood from easily collectible data has several important implications for the early detection, intervention, and treatment of mental health disorders. One promising data source to help monitor human behavior is daily smartphone usage. However, care must be taken to summarize behaviors without identifying the user through personal (e.g., personally identifiable information) or protected (e.g., race, gender) attributes. In this paper, we study behavioral markers of daily mood using a recent dataset of mobile behaviors from adolescent populations at high risk of suicidal behaviors. Using computational models, we find that language and multimodal representations of mobile \textit{typed text} (spanning typed characters, words, keystroke timings, and app usage) are predictive of daily mood. However, we find that models trained to predict mood often also capture private user identities in their intermediate representations. To tackle this problem, we evaluate approaches that obfuscate user identity while remaining predictive. By combining multimodal representations with privacy-preserving learning, we are able to push forward the performance-privacy frontier.\blfootnote{$^\star$first two authors contributed equally.}
\end{abstract}

\vspace{-1mm}
\section{Introduction}
\vspace{-1mm}

\begin{figure}%
    \centering
    \vspace{-0mm}
    \includegraphics[width=\linewidth]{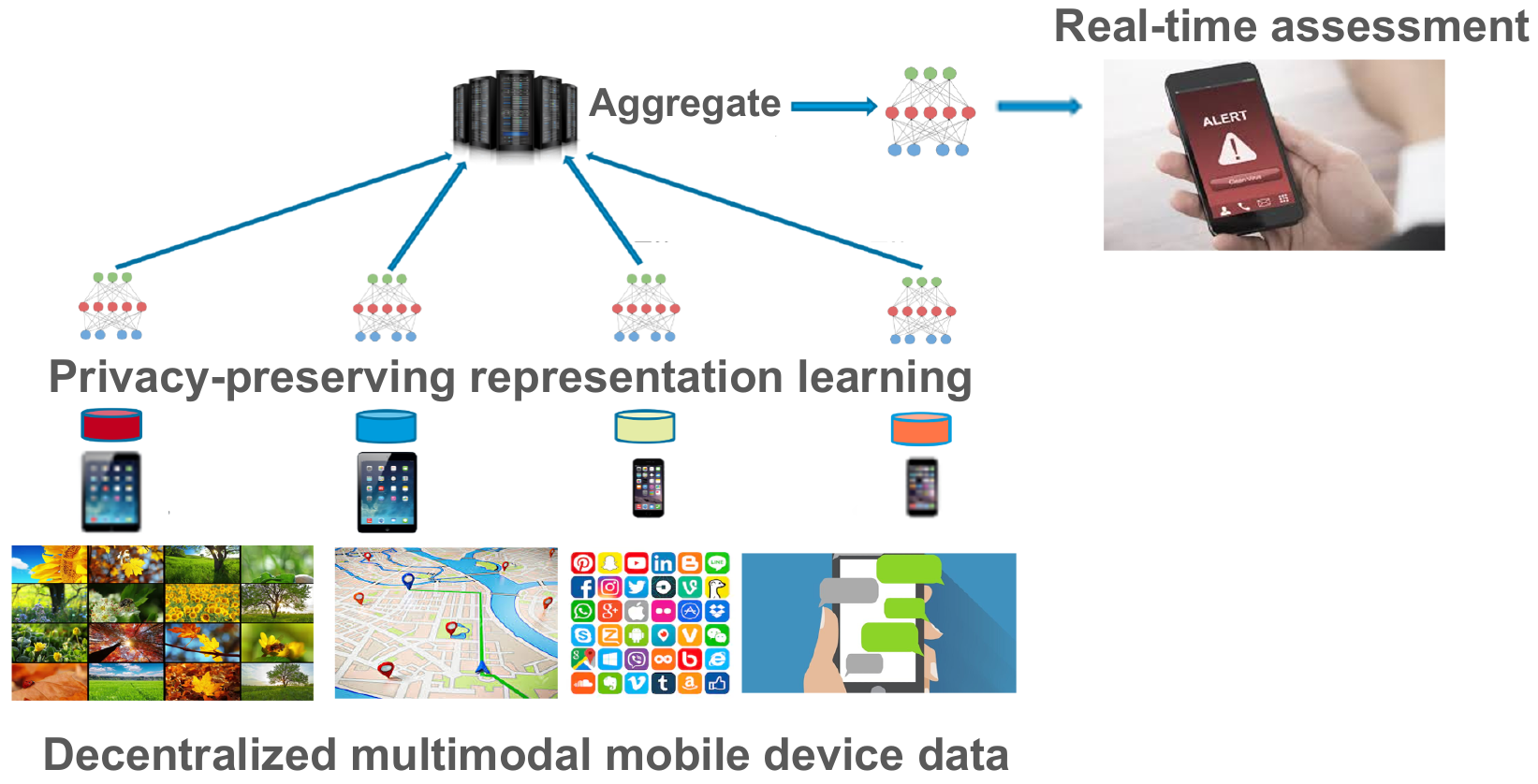}
    \vspace{-4mm}
    \caption{Intensive monitoring of behaviors via adolescents' natural use of smartphones may help identify real-time predictors of mood in high-risk youth as a proxy for suicide risk. While smartphones provide a valuable data source spanning text, keystrokes, app usage, and geolocation, one must take care to summarize behaviors without revealing user identities through personal (e.g., personally identifiable information) or protected attributes (e.g., race, gender) to potentially adversarial third parties.\vspace{-4mm}}
    \label{fig:overview}
\end{figure}

Mental illnesses can have a damaging permanent impact on communities, societies, and economies all over the world~\citep{world2003investing}. Individuals often do not realize they are at risk of mental disorders even when they have symptoms. As a result, many are late in seeking professional help and treatment~\citep{thornicroft2016evidence}, particularly among adolescents where suicide is the second leading cause of death~\citep{curtin2019death}. In addition to deaths, $16\%$ of high school students report having serious suicidal thoughts each year, and $8\%$ of them make one or more suicide attempts~\citep{cdc}. This problem is particularly exacerbated as an ``echo pandemic'' of mental health problems have arisen in the wake of the COVID-$19$ pandemic~\cite{inkster2021early,saha2020psychosocial}.

Intensive monitoring of behaviors via adolescents' natural use of smartphones may help identify real-time predictors of mood in high-risk youth as a \textit{proxy} for suicide risk~\citep{nahum2018just}. While there are inherent limitations in the mismatch between mood prediction and ultimately developing real-time intervention against imminent suicide risk~\cite{coppersmith2018natural,ophir2020deep}, we believe that the former is a reasonable starting point to tackle similar machine learning problems surrounding affective computing and privacy-preserving learning. Studying mood in this high-risk population is a valuable goal given that suicide attempts are often decided within a short time-lapse and just-in-time assessments of mood changes can be a stepping stone in this direction~\cite{rizk2019variability,oquendo2020highly}. Technologies for mood prediction can also be a valuable component of decision support for clinicians and healthcare providers during their assessments~\cite{mann2006can,cho2019mood}.

\textbf{Recent work} in affective computing has begun to explore the potential in predicting mood from mobile data. Studies have found that typing patterns~\citep{cao2017deepmood,ghosh2017evaluating,huang2018dpmood,zulueta2018predicting}, self-reporting apps~\citep{suhara2017deepmood}, and wearable sensors~\citep{ghosh2017tapsense,sano2018identifying} are particularly predictive. In addition, multimodal modeling of multiple sensors (e.g., wearable sensors and smartphone apps) was shown to further improve performance~\citep{jaques2017multimodal,taylor2017personalized}. While current work primarily relies on self-report apps for \textit{long-term} mood assessments~\citep{glenn2014improving}, our work investigates mobile behaviors from a high-risk teenage population as a predictive signal for \textit{daily} mood~\citep{franklin2017risk,large2017patient}.

Prior work has also shown that private information is predictable from digital records of human behavior~\citep{kosinski2013private}, which is dangerous especially when sensitive user data is involved. As a result, in parallel to improving predictive performance, a recent focus has been on improving privacy through techniques such as differential privacy~\citep{dankar2012application,dankar2013practicing,dankar2012estimating} and federated learning~\citep{DBLP:journals/corr/McMahanMRA16,geyer2017differentially,liang2020think}, especially for healthcare data (e.g., electronic health records~\citep{xu2019federated}) and wearable devices~\citep{chen2020fedhealth}.

\textbf{In this paper}, as a step towards using \textit{multimodal privacy-preserving} mood prediction as fine-grained signals to aid in mental health assessment, we analyze a recent dataset of mobile behaviors collected from adolescent populations at high suicidal risk.
With consent from participating groups, the dataset collects fine-grained features spanning online communication, keystroke patterns, and application usage. Participants are administered daily questions probing for mood scores.
By collecting and working on ground-truth data for this population, we are able to benchmark on a more accurate indicator of mood rather than proxy data such as mood signals inferred from social media content or behavior~\cite{ernala2019methodological}.
This unique dataset presents an opportunity to investigate a different medium of natural language processing - \textit{typed text} which presents new challenges beyond conventionally studied written~\citep{marcus1993building} and spoken~\citep{marslen1980temporal} text. We propose multimodal models that \textit{contextualize} text with their typing speeds and app usage. However, these models often capture private user identities in their intermediate representations when predicting mood. As a step towards privacy-preserving learning, we also propose approaches that obfuscate user identity while remaining predictive of daily mood. By combining multimodal contextualization with privacy-preserving learning, we are able to push forward the performance-privacy frontier. Finally, we conclude with several observations regarding the uniqueness of typed text as an opportunity for NLP on mobile data.

\vspace{-1mm}
\section{Multimodal Mobile Dataset}
\label{dataset}
\vspace{-1mm}

Intensive monitoring of behaviors via adolescents' frequent use of smartphones may shed new light on the early risk of suicidal thoughts and ideations~\citep{nahum2018just}. Smartphones provide a valuable and natural data source with rich behavioral markers spanning online communication, keystroke patterns, and application usage. Learning these markers requires large datasets with diversity in participants, variety in features, and accuracy in annotations. As a step towards this goal, we recently collected a dataset of mobile behaviors from high-risk adolescent populations with consent from participating groups.

We begin with a brief review of the data collection process. This data monitors adolescents spanning (a) recent suicide attempters (past 6 months) with current suicidal ideation, (b) suicide ideators with no past suicide attempts, and (c) psychiatric controls with no history of suicide ideation or attempts. Passive sensing data is collected from each participant’s smartphone across a duration of $6$ months. Participants are administered clinical interviews probing for suicidal thoughts and behaviors (STBs), and self-report instruments regarding symptoms and acute events (e.g., suicide attempts, psychiatric hospitalizations) are tracked weekly via a questionnaire. All users have given consent for their mobile data to be collected and shared with us for research purposes. This study has been carefully reviewed and approved by an IRB. We follow the NIH guidelines, with a central IRB (single IRB) linked to secondary sites. We have IRB approval for the central institution and all secondary sites.

\vspace{-1mm}
\subsection{Mood Assessment via Self-Report}
\vspace{-1mm}

Every day at $8$am, users are asked to respond to the following question - ``In general, how have you been feeling over the last day?'' - with an integer score between $0$ and $100$, where $0$ means very negative and $100$ means very positive. To construct our prediction task, we discretized these scores into the following three bins: \textit{negative} ($0-33$), \textit{neutral} ($34-66$), and \textit{positive} ($67-100$), which follow a class distribution of $12.43\%$, $43.63\%$, and $43.94\%$ respectively. For our $3$-way classification task, participants with fewer than $50$ daily self-reports were removed since these participants do not provide enough data to train an effective model. In total, our dataset consists of $1641$ samples, consisting of data coming from $17$ unique participants.

\vspace{-1mm}
\subsection{Features}
\vspace{-1mm}

We focused on keyboard data, which includes the time of data capture, the mobile application used, and the text entered by the user. For each daily score response at $8$am, we use information collected between $5$am on the previous day to $5$am on the current day. We chose this $5$am-$5$am window by looking at mobile activity and finding the lowest activity point when most people ended their day: $5$am. Since users report the \textit{previous} day’s mood (when prompted at $8$am), we decided to use this $5$am-$5$am time period to summarize the previous day’s activities. Through prototyping, this prompt time and frequency were found to give reliable indicators of the previous day’s mood. From this window, we extracted the following features to characterize and contextualize typed text.

\vspace{-1mm}
\textit{Text}: After removing stop-words, we collected the top $1000$ words (out of approximately $3.2$ million) used across all users in our dataset and created a \textit{bag-of-words} feature that contains the daily number of occurrences of each word.

\vspace{-1mm}
\textit{Keystrokes}: We also extracted keystroke features that record the exact timing that each character was typed on a mobile keyboard (including alphanumeric characters, special characters, spaces, backspace, enter, and autocorrect). By taking the increase in recorded timing after each keystroke, we obtain the duration that each key was pressed in a sequence of keystrokes during the day. When extracting keystrokes, we removed all small timings under $10^{-2}$ seconds.

\vspace{-1mm}
\textit{App usage}: We count the number of mobile applications used per day, creating a \textit{bag-of-apps} feature for each day. We discard applications that are used by less than $10\%$ of the participants so that our features are generalizable to more than just a single user in the dataset, resulting in $137$ total apps (out of the original $640$).

In a preliminary analysis, we observed that predictive models performed well when binarizing our feature vectors into boolean vectors, which signify whether a word or app was used on a given day (i.e., mapping values greater than $0$ to $1$). Our final feature vectors consist of a concatenation of a normalized and a binarized feature vector, resulting in $2000$ and $274$-dimensional vectors for text and app features respectively. For keystrokes, we found that summarizing the sequence of timings using a histogram (i.e., defining a set of timing buckets and creating a \textit{bag-of-timings} feature) for each day performed well. We chose $100$ fine-grained buckets, resulting in a $100$-dimensional keystroke vector. Please refer to Appendix~\ref{dataset_details} for additional details about the dataset and extracted features.

\begin{figure*}%
    \centering
    \vspace{-0mm}
    \includegraphics[width=0.9\textwidth]{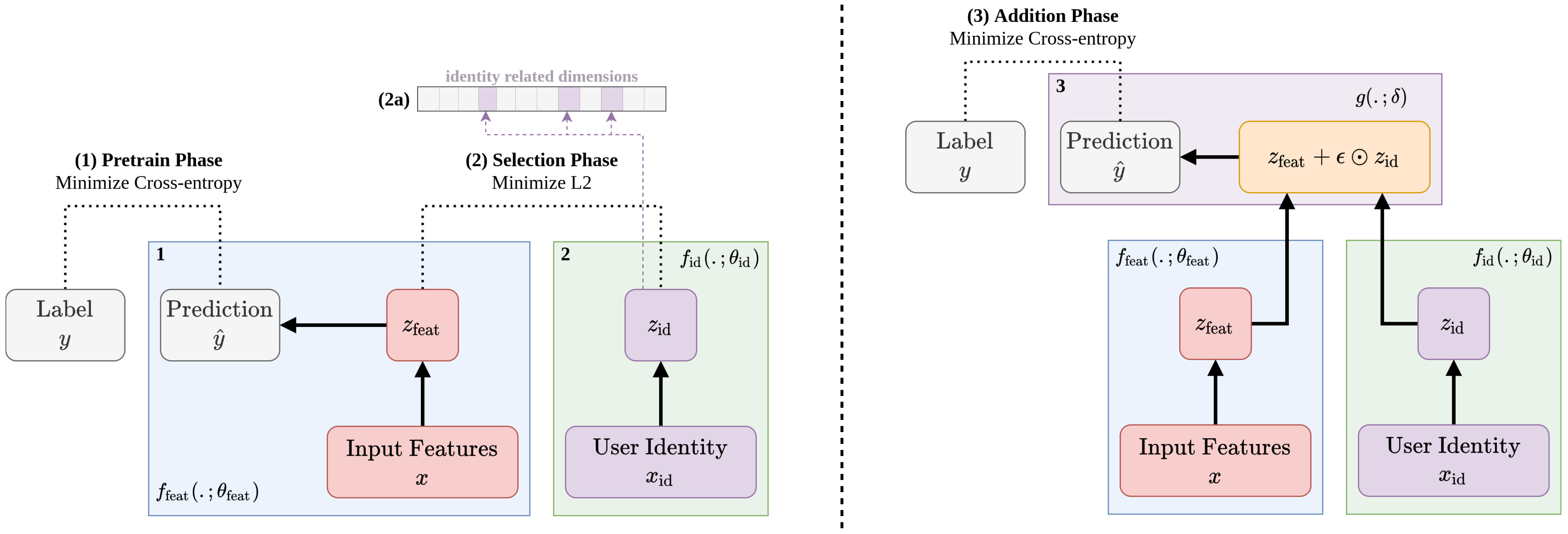}
    \vspace{-2mm}
    \caption{Diagram of the \names\ algorithm learned via the (1) \textit{pretrain}, (2) \textit{selection}, and (3) \textit{addition} phases. Boxes with numbers denote which parameters are being optimized in the corresponding step. For example, in the \textit{addition} phase (3), \names\ optimizes parameters $\delta$ in $g(.;\delta)$. (2a) depicts identity-dependent dimensions $z_\textrm{id}$, which is a sparse vector of size $\text{dim} (z_\textrm{feat})$ whose nonzero values (colored purple) signify dimensions of the identity-dependent subspace in $z_\textrm{feat}$.\vspace{-4mm}}%
    \label{fig:sal}%
\end{figure*}

\vspace{-1mm}
\section{Mood Prediction Methods}
\vspace{-1mm}

In this paper, we focus on studying approaches for learning privacy-preserving representations from mobile data for mood prediction. Our processed data comes in the form of $\{(x_{t,i}, x_{k,i}, x_{a,i}, y_i)\}_{i=1}^n$ with $x_t \in \mathbb{N}^{|V_t|=2000}$ denoting the bag-of-words features, $x_k \in \mathbb{N}^{|V_k|=100}$ denoting the bag-of-timings features, and $x_a \in \mathbb{N}^{|V_a|=274}$ denoting the bag-of-apps features. $y$ denotes the label which takes on one of our $3$ mood categories: negative, neutral, and positive. In parallel, we also have data representing the corresponding (one-hot) user identity $x_{\textrm{id}}$ which will be useful when learning privacy-preserving representations that do not encode information about user identity $x_{\textrm{id}}$ and evaluating privacy performance.

\vspace{-1mm}
\subsection{Unimodal Approaches}
\vspace{-1mm}

We considered two unimodal baselines:

\vspace{-1mm}
1. Support Vector Machines (\textsc{SVMs}) project training examples to a chosen kernel space and finds the optimal hyperplane that maximally separates each class of instances. We apply an SVM classifier on input data $x_{\textrm{uni}} \in \{ x_t, x_k, x_a \}$ and use supervised learning to predict daily mood labels $y$.

\vspace{-1mm}
2. Multilayer Perceptrons (\textsc{MLPs}) have seen widespread success in supervised prediction tasks due to their ability in modeling complex nonlinear relationships. Because of the small size of our dataset, we choose a simple multilayer perceptron with two hidden layers. Similarly, we apply an \textsc{MLP} classifier on input data $x_{\textrm{uni}} \in \{ x_t, x_k, x_a \}$ to predict daily mood labels $y$.

\vspace{-1mm}
\subsection{Multimodal Models}
\vspace{-1mm}

We extend both \textsc{SVM} and \textsc{MLP} classifiers using early fusion~\citep{baltruvsaitis2018multimodal} of text and app usage to model multimodal interactions. Specifically, we align the input through concatenating the bag-of-words, bag-of-keystrokes, and bag-of-apps features for each day resulting in an input vector $x_{\textrm{multi}} = x_t \oplus x_k \oplus x_a$, before using an \textsc{SVM}/\textsc{MLP} classifier for prediction.

\vspace{-1mm}
\subsection{A Step Toward Preserving Privacy}
\vspace{-1mm}

While classifiers trained with traditional supervised learning can learn useful representations for mood prediction, they carry the risk of \textit{memorizing} the identity of the user along with their sensitive mobile usage and baseline mood scores, and possibly \textit{revealing} these identities to adversarial third-parties~\citep{abadi2016deep}. Therefore, it is crucial to perform mood prediction while also protecting the privacy of personal identities.

We adapt the Selective-Additive Learning (SAL) framework~\citep{wang2017select} for the purpose of privacy-preserving learning. While SAL was originally developed with a very different goal in mind: improving model generalization, we expand SAL to a very important problem in healthcare: preserving privacy. We adapted SAL to learn \textit{disentangled} representations separated into \textit{identity-dependent} private information and \textit{identity-independent} population-level information using three phases:

\vspace{-1mm}
(1) \textit{Pretrain phase:} The input is a set of (multimodal) features $x$ that are likely to contain both identity-dependent and independent information. The intermediate representation $z_\textrm{feat} = f_\textrm{feat}(x; \theta_\textrm{feat}^*)$ is obtained from an \textsc{MLP} classifier pretrained for mood prediction. $f_\textrm{feat}$ denotes the classifier with pretrained parameters $\theta_\textrm{feat}^*$.

\definecolor{gg}{RGB}{15,125,15}
\definecolor{rr}{RGB}{190,45,45}

\begin{table*}[!tbp]
\fontsize{8.5}{11}\selectfont
\centering
\caption{Comparison of mood prediction performance across different modalities. Best results in \textbf{bold}. For both accuracy and F1 score, models jointly trained on text, keystroke, and apps features outperform models trained using individual modalities. $\star$ denotes that the difference between multimodal and all unimodal models is statistically significant (p-value $<< 0.05$).\vspace{-2mm}}
\setlength\tabcolsep{5.0pt}
\begin{tabular}{l || c c c c | c c c c }
\Xhline{3\arrayrulewidth}
& \multicolumn{4}{c|}{\textsc{F1 score}} & \multicolumn{4}{c}{\textsc{Accuracy}} \\
Modalities & \textsc{Baseline} & \textsc{SVM} & \textsc{MLP} & \names &  \textsc{Baseline} & \textsc{SVM} & \textsc{MLP} & \names \\
\Xhline{3\arrayrulewidth}
Text + Keystrokes + Apps & $19.07$ & $\textbf{62.81}^{\star}$ & $\textbf{59.61}^{\star}$ & $\textbf{60.11}^{\star}$ & $40.18$ & $\textbf{67.43}^{\star}$ & $\textbf{63.59}^{\star}$ & $\textbf{64.06}^{\star}$\\
Text + Keystrokes & $19.07$ & $61.19$ & $57.65$ & $58.70$ & $40.18$ & $65.87$ & $61.81$ & $62.61$ \\
Text + Apps &  $19.07$ & $62.08$ & $58.38$ & $52.90$ & $40.18$ & $66.59$ & $62.93$ & $56.76$ \\
\Xhline{0.5\arrayrulewidth}
Text & $19.07$ & $61.15 $ & $56.27$ & $52.63$ & $40.18$ & $65.83$ & $60.61$ & $56.08$ \\
Keystrokes & $19.07$ & $57.68$ & $51.43$ & $34.73$ & $40.18$ & $61.03$ & $55.87$
& $39.18$ \\
Apps & $19.07$ & $58.65$ & $52.29$ & $51.32$ & $40.18$ & $62.65$ & $55.26$ & $55.68$ \\
\Xhline{3\arrayrulewidth}
\end{tabular}
\vspace{-4mm}
\label{modalities_comparison}
\end{table*}

\vspace{-1mm}
(2) \textit{Selection phase:} Our goal is to now disentangle the identity-dependent and independent information within $z_\textrm{feat}$. We hypothesize that dependent and independent information are encoded in separate subspaces of the feature vector $z_\textrm{feat}$. This allows us to disentangle them by training a separate classifier to predict $z_\textrm{feat}$ \textit{as much as possible} given only the user identity:
\begin{equation}
\label{eq:SAL}
    \theta_\textrm{id}^* = \argmin_{\theta_\textrm{id}} \left( z_\textrm{feat} - f_\textrm{id} (x_\textrm{id}; \theta_\textrm{id}) \right)^2 + \lambda ||z_{\textrm{id}}||_1,
\end{equation}
where $x_\textrm{id}$ denotes a one hot encoding of user identity as input, $f_\textrm{id}$ denotes the identity encoder with parameters $\theta_\textrm{id}$, and $\lambda$ denotes a hyperparameter that controls the weight of the $\ell_1$ regularizer. $f_\textrm{id}$ projects the user identity encodings to the feature space learned by $f_\textrm{feat}$. By minimizing the objective in equation~\eqref{eq:SAL} for each $(x, x_\textrm{id})$ pair, $f_\textrm{id}$ learns to encode user identity into a sparse vector $z_\textrm{id} = f_\textrm{id} (x_\textrm{id}; \theta_\textrm{id}^*)$ representing identity-dependent features: the nonzero values of $z_\textrm{id}$ represent dimensions of the identity-dependent subspace in $z_\textrm{feat}$, while the remaining dimensions belong to the identity-independent subspace.

\vspace{-1mm}
(3) \textit{Addition phase:} Given two factors $z_\textrm{feat}$ and $z_\textrm{id}$, to ensure that our prediction model does not capture identity-related information $z_\textrm{id}$, we add multiplicative Gaussian noise to remove information from the identity-related subspace $z_\textrm{id}$ while repeatedly optimizing for mood prediction with a final \textsc{MLP} classification layer $g(z_\textrm{feat}, z_\textrm{id} ; \delta)$. This resulting model should only retain identity-independent features for mood prediction:
\begin{equation}
    \hat{y} = g \left( z_\textrm{feat} + \epsilon \odot z_\textrm{id} \right)
\end{equation}
where $\epsilon \sim N(0, \sigma^2)$ is repeatedly sampled across batches and training epochs. We call this approach \namel, or \names\ for short, and summarize the final algorithm in Figure~\ref{fig:sal}.

\textbf{Controlling the tradeoff between performance and privacy:} There is often a tradeoff between privacy and prediction performance. To control this tradeoff, we vary the parameter $\sigma$, which is the variance of noise added to the identity-dependent subspace across batches and training epochs. $\sigma=0$ recovers a standard \textsc{MLP} with good performance but reveals user identities, while large $\sigma$ effectively protects user identities but at the possible expense of mood prediction performance. In practice, the optimal tradeoff between privacy and performance varies depending on the problem. For our purposes, we automatically perform model selection using this performance-privacy ratio $R$ computed on the validation set, where
\begin{equation}
    R = \frac{s_{\textsc{MLP}} - s_{\names}}{t_{\textsc{MLP}} - t_{\names}}    
\end{equation}
is defined as the improvement in privacy per unit of performance lost. Here, $s$ is defined as the accuracy in user prediction and $t$ is defined as the F1 score on mood prediction.

\vspace{-1mm}
\section{Experiments}
\vspace{-1mm}

We perform experiments to test the utility of text, keystroke, and app features in predicting daily mood while keeping user privacy in mind.

\vspace{-1mm}
\subsection{Experimental Setup}
\vspace{-1mm}

\textit{Data splits:} Given that our data is longitudinal, we split our data into 10 partitions ordered chronologically by users. We do so in order to maintain independence between the train, validation, and test splits in the case where there is some form of time-level dependency within our labels.

\textit{Evaluation:} For each model, we run a nested $k$-fold cross-validation (i.e., we perform $9$-fold validation within $10$-fold testing). For each test fold, we identify the optimal parameter set as the one that achieves the highest mean validation score over the validation folds. To evaluate \names, we use the best performing \textsc{MLP} model for each test fold as our base classifier before performing privacy-preserving learning. For all experiments, we report the test accuracy and macro F1 score because our classes are imbalanced. Given the low number of cross-validation folds, we use the Wilcoxon signed-rank test~\cite{wilcoxon1992individual} at $5\%$ significance level for all statistical comparisons (see Appendix~\ref{exp_details} for more experimental details).

\vspace{-1mm}
\subsection{Results on Mood Prediction}
\vspace{-1mm}

We make the following observations regarding the learned language and multimodal representations for mood prediction:

\vspace{-1mm}
\textbf{Observation 1: Text, keystroke, and app usage features are individually predictive of mood.} To evaluate how predictive our extracted text, keystroke timings, and app usage features are, we first run experiments using \textsc{SVM}, \textsc{MLP}, and \names\ on each individual feature separately. Since we have unbalanced classes, we chose a majority classifier (i.e., most common class in the training set) as our baseline. From Table~\ref{modalities_comparison}, we observe that using these three feature types individually outperforms the baseline with respect to accuracy and F1 score. Using the Wilcoxon signed-rank test~\cite{wilcoxon1992individual} at $5\%$ significance level, we found that these improvements over the baseline in both F1 score and accuracy are statistically significant (p-value $<< 0.05$).

\definecolor{gg}{RGB}{15,125,15}
\definecolor{rr}{RGB}{190,45,45}

\begin{table}[!tbp]
\fontsize{8.5}{11}\selectfont
\centering
\caption{Mood prediction from text using extended pretrained LM encoders. We find that these models struggle on extremely long contexts of typed text.\vspace{-2mm}}
\setlength\tabcolsep{5.0pt}
\begin{tabular}{l || c | c}
\Xhline{3\arrayrulewidth}
Models & \textsc{F1 score} & \textsc{Accuracy} \\
\Xhline{3\arrayrulewidth}
BoW & $\mathbf{56.27}$ & $\mathbf{60.61}$ \\
BERT & $51.42$ & $58.06$ \\
XLNet & $19.85$ & $42.40$ \\
LongFormer & $19.85$ & $42.40$ \\
\Xhline{3\arrayrulewidth}
\end{tabular}
\vspace{-4mm}
\label{text_lms}
\end{table}

\vspace{-1mm}
\textbf{Observation 2: Pretrained sentence encoders struggle on this task.} We also applied pretrained sentence encoders such as BERT~\citep{devlin2019bert} on the language modality for mood prediction. Surprisingly, we found that none of these approaches performed stronger than a simple bag-of-words (see Table~\ref{text_lms}). We provide two possible explanations for this phenomenon:

\vspace{-1mm}
1. BERT is suitable for written text on the web (Wikipedia, BookCorpus, carefully human-annotated datasets) which may not generalize to informal typed text that contains emojis, typos, and abbreviations (see Section~\ref{qualitative} for a qualitative analysis regarding the predictive abilities of emojis and keystrokes for mood prediction).

\vspace{-1mm}
2. We hypothesize that it is difficult to capture such long sequences of data (>$1000$ time steps) spread out over a day. Current work has shown that BERT struggles with long sequence lengths~\citep{beltagy2020longformer}. We trained two extensions XLNet~\citep{yang2019xlnet} and LongFormer~\citep{beltagy2020longformer} specifically designed to take in long-range context but found that they still underperform as compared to a simple bag-of-words approach.

\vspace{-1mm}
\textbf{Observation 3: Fusing both text and keystroke timings improves performance.} This dataset presents a unique opportunity to study representations of \textit{typed} text as an alternative to conventionally studied written or spoken text. While the latter two use language alone, typed text includes keystroke features providing information about the timings of when each character was typed. In Table~\ref{modalities_comparison}, we present some of our initial results in learning text and keystroke representations for mood prediction and show consistent improvements over text alone. We further study the uniqueness of typed text by comparing the following baselines:

\vspace{-1mm}
1. \textit{Text}: bag-of-words only.

\vspace{-1mm}
2. \textit{Text + char keystrokes}: bag-of-words and bag-of-timings across all characters.

\vspace{-1mm}
3. \textit{Text + split char keystrokes}: bag-of-words and bag-of-timings subdivided between $6$ groups: alphanumeric characters, symbols, spacebar, enter, delete, and use of autocorrect. This baseline presents a more fine-grained decomposition of the typing speeds across different semantically related character groups.

\vspace{-1mm}
4. \textit{Text + word keystrokes}: bag-of-words and bag-of-timings summed up over the characters in each word. This presents a more interpretable model to analyze the relationships between words and the distribution of their typing speeds.

\definecolor{gg}{RGB}{15,125,15}
\definecolor{rr}{RGB}{190,45,45}

\begin{table}[!tbp]
\fontsize{8.5}{11}\selectfont
\centering
\caption{Mood prediction using a \textsc{MLP} from text and keystroke features tallied from (1) all characters, (2) a split between types of characters, as well as (3) aggregated across words.\vspace{-2mm}}
\setlength\tabcolsep{5.0pt}
\begin{tabular}{l || c | c c c | c c c | c c c}
\Xhline{3\arrayrulewidth}
Modalities & \textsc{F1 score} & \textsc{Accuracy} \\
\Xhline{3\arrayrulewidth}
Text & $56.27$ & $60.61$ \\
Text + Char keystrokes & $\mathbf{57.65}$ & $\mathbf{61.81}$ \\
Text + Split char keystrokes & $57.32$ & $61.21$ \\
Text + Word keystrokes & $56.46$ & $60.68$ \\
\Xhline{3\arrayrulewidth}
\end{tabular}
\vspace{-4mm}
\label{text_keystrokes}
\end{table}

From Table~\ref{text_keystrokes}, we observe that keystrokes accurately contextualize text, especially when using fine-grained keystroke distributions across individual characters. Other methods incorporating keystroke features are also all stronger than unimodal models. Different ways of representing keystrokes also provide different levels of interpretability regarding the relationships between words, characters, and keystrokes for mood prediction, which we qualitatively analyze in \S\ref{qualitative}.

\begin{figure*}%
    \centering
    \vspace{-0mm}
    \subfloat[\centering \textsc{MLP} (without privacy-preserving)]{{\includegraphics[width=5.5cm]{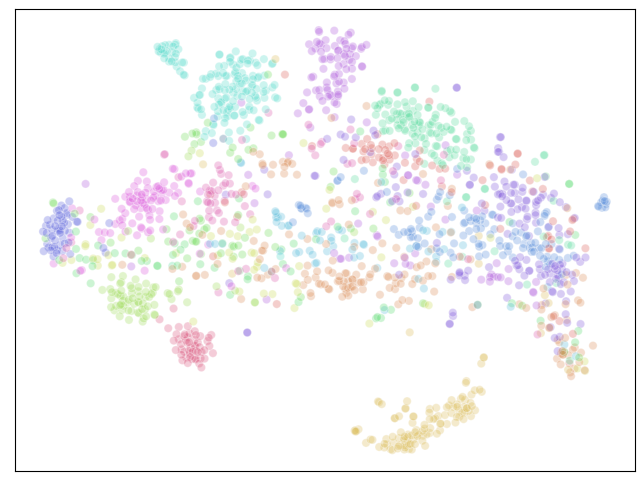} }}%
    \qquad \qquad \qquad \quad
    \subfloat[\centering \names\ (with privacy-preserving)]{{\includegraphics[width=5.5cm]{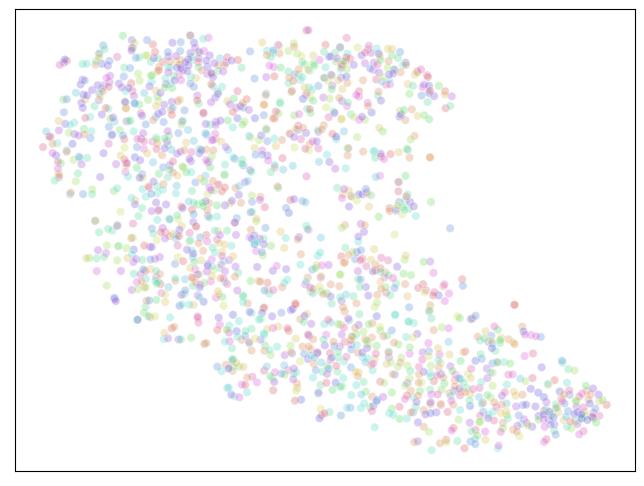} }}%
    \vspace{-2mm}
    \caption{Visualization of representations learned by (a) \textsc{MLP} and (b) \names, which have been reduced to two dimensions via t-SNE and colored by participant identity. Representations learned by \names\ are no longer separable by users which better preserves privacy.\vspace{-4mm}}
    \label{fig:privacy_representations}
\end{figure*}

\vspace{-1mm}
\textbf{Observation 4: Multimodal representation learning achieves the best performance.} In Table \ref{modalities_comparison}, we also compare the performance of our models on combined (text + keystroke + apps) features versus the performance on each individual feature set. For both metrics, combining all features gives better performance over either subset.

\vspace{-1mm}
\subsection{Results on Preserving Privacy}
\vspace{-1mm}

Despite these promising results in mood prediction, we ask an important question: \textit{Does the model capture user identities as an intermediate step towards predicting mood?} To answer this question, we analyze the privacy of raw mobile data and trained models. We then study our proposed method of learning privacy-preserving features to determine whether it can obfuscate user identity while remaining predictive of daily mood.

\vspace{-1mm}
\textbf{How private is the mobile data?} We evaluate how much the data reveal user identities by training predictive models with typed text, keystroke timings, and app usage as input and user identity as the prediction target. From Table~\ref{privacy_data}, we observe that all modalities are very predictive of user identity (>$87\%$ accuracy), which further motivates the need to learn privacy-preserving features. We further note that identifiable information can be very subtle: while only $28/1000$ words were named entities, it was possible to identify the user identity with >$87\%$ accuracy, which means that subtle word choice can be identify the user (similarly for apps and keystrokes).

\vspace{-1mm}
\textbf{How private are the learned privacy-preserving features?} We also study whether our learned features are correlated with user identity through both visualizations and quantitative evaluations.

\definecolor{gg}{RGB}{15,125,15}
\definecolor{rr}{RGB}{190,45,45}

\begin{table}[!tbp]
\fontsize{8.5}{11}\selectfont
\centering
\caption{We report user identity prediction performance from raw input data and find that identities are \textbf{very easily revealed} from text, keystrokes, and app usage.\vspace{-2mm}}
\setlength\tabcolsep{5.0pt}
\begin{tabular}{l || c c | c c }
\Xhline{3\arrayrulewidth}
& \multicolumn{2}{c|}{\textsc{F1 score}} & \multicolumn{2}{c}{\textsc{Accuracy}} \\
Modalities & \textsc{SVM} & \textsc{MLP} & \textsc{SVM} & \textsc{MLP} \\
\Xhline{3\arrayrulewidth}
Text & $89.42$ & $92.05$ & $90.60$ & $93.12$ \\
Keystrokes & $91.36$ & $87.04$ & $90.98$ & $87.15$ \\
Apps & $85.68$ & $87.49$ & $90.91$ & $92.00$ \\
\Xhline{3\arrayrulewidth}
\end{tabular}
\vspace{-4mm}
\label{privacy_data}
\end{table}

\vspace{-1mm}
\textit{Visualizations:} We use t-SNE~\citep{van2008visualizing} to reduce the learned features from trained models to $2$ dimensions. After color-coding the points by participant identity, we identify distinct clusters in Figure~\ref{fig:privacy_representations}(a), which implies that mood prediction can be strongly linked to identifying the person, therefore coming at the price of losing privacy.

As an attempt to reduce reliance on user identity, we train \names\ which is designed to obfuscate user-dependent features. After training \names, we again visualize the representations learned in Figure \ref{fig:privacy_representations}(b) and we find that they are less visually separable by users, indicating that \names\ indeed learns more user-independent features.

\definecolor{gg}{RGB}{15,125,15}
\definecolor{rr}{RGB}{190,45,45}

\begin{table}[t!]
\fontsize{8.5}{11}\selectfont
\centering
\caption{Comparison of our privacy-preserving approach (\names) with the baseline (\textsc{MLP}). We evaluate privacy in predicting user identity from learned \textbf{representations} (\textbf{lower} accuracy is better), and find that \names\ effectively obfuscates user identity while retaining performance. T: text, K: keystrokes, A: apps.\vspace{-2mm}}
\setlength\tabcolsep{6.0pt}
\begin{tabular}{l || c c | c c}
\Xhline{3\arrayrulewidth}
& \multicolumn{2}{c|}{\textsc{Performance} ($\uparrow$)} & \multicolumn{2}{c}{\textsc{Privacy} ($\downarrow$)}  \\
Modalities & \textsc{MLP} & \names & \textsc{MLP} & \names \\
\Xhline{3\arrayrulewidth}
T + K + A & $59.61$ & $58.48$ & $71.47$ & $\mathbf{34.49}$\\
T + K & $57.65$ & $57.40$ & $64.17$ & $\mathbf{30.99}$ \\
T + A & $58.38$ & $57.76$ & $79.04$ & $\mathbf{65.13}$ \\
T & $56.27$ & $54.11$ & $76.41$ & $\mathbf{52.20}$ \\
K & $51.43$ & $42.48$ & $55.61$ & $\mathbf{25.71}$ \\
A & $52.29$ & $49.15$ & $85.94$ & $\mathbf{66.74}$ \\
\Xhline{3\arrayrulewidth}
\end{tabular}
\label{privacy_features}
\vspace{-4mm}
\end{table}

\vspace{-1mm}
\textit{Quantitative evaluation:} To empirically evaluate how well our models preserve privacy, we extracted the final layer of each trained model and fit a logistic regression model to predict user identity using these final layer representations as input. The more a model preserves privacy, the harder it should be to predict user identity. From Table~\ref{privacy_features}, we observe that we can predict user identity based on the learned \textsc{MLP} representations with high accuracy (>$85\%$) using the most sensitive app usage features. For other modality combinations, user identity can also be decoded with more than $70\%$ accuracy with the exception of keystrokes which are the most private ($55\%$). We achieve significantly more privacy using \names\ embeddings - roughly $35\%$ for the best multimodal model, which indicates the possibility of \names\ as a means of achieving privacy-preserving mood prediction.

\begin{figure}%
    \centering
    \vspace{-0mm}
    \includegraphics[width=\linewidth]{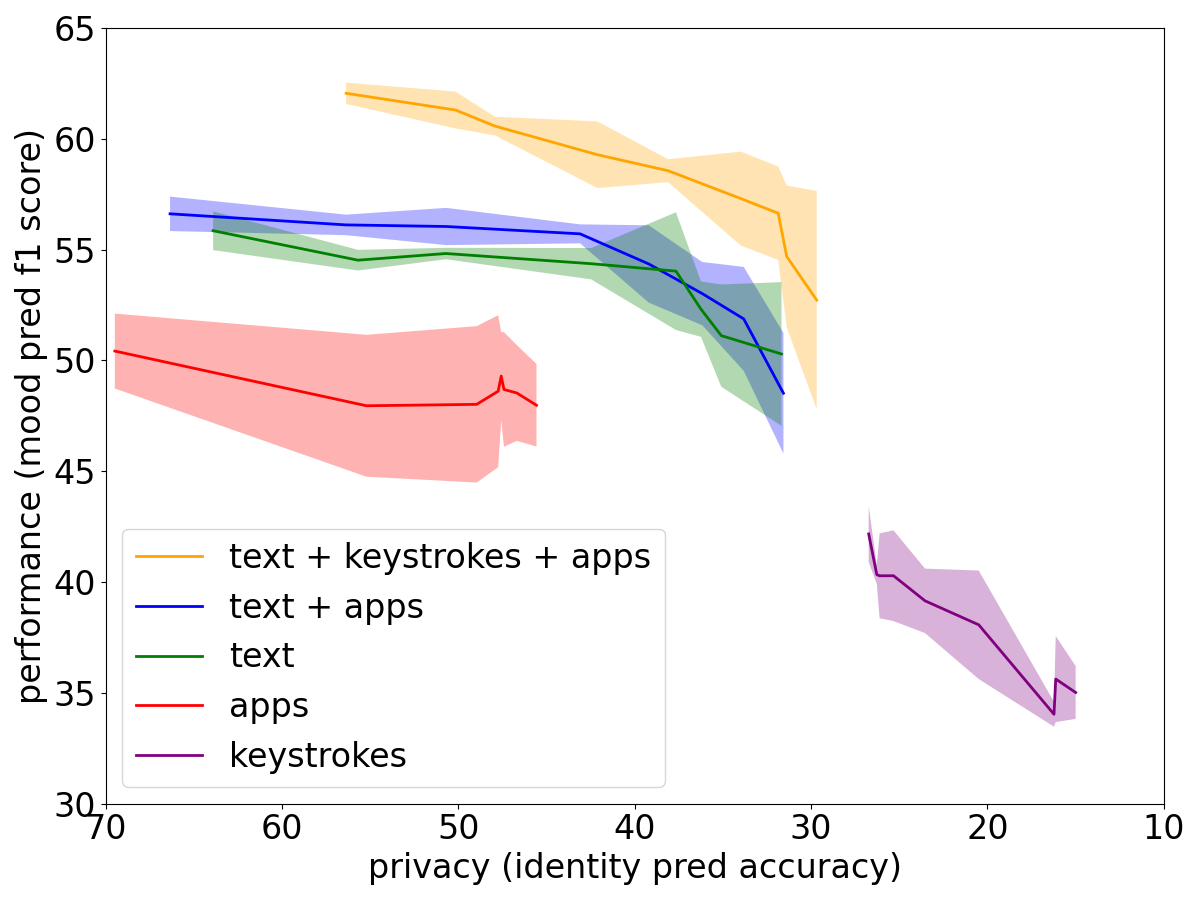}
    \vspace{-4mm}
    \caption{Tradeoff between performance (mood prediction F1 score, \textbf{higher} is better) and privacy (identity prediction accuracy, \textbf{lower} is better). Shaded regions denote standard deviations from the mean (solid lines). \names\ provides a tunable parameter $\sigma$ to control the tradeoff, which allows us to plot a range of (performance, privacy) points. Using a multimodal model on text, keystroke, and app features obtains better performance and privacy at the same time.\vspace{-4mm}}
    \label{fig:tradeoff}%
\end{figure}

\vspace{-1mm}
\textbf{Understanding the tradeoff between performance and privacy:} \names\ provides a tunable parameter $\sigma$ to control the variance of noise applied on the identity-related dimensions. This parameter $\sigma$ has the potential to give a tradeoff between privacy and prediction performance. In Figure~\ref{fig:tradeoff}, we plot this tradeoff between performance (mood prediction F1 score, higher is better) and privacy (identity prediction accuracy, lower is better). We find that keystroke features, while themselves not very useful in predicting mood, are highly private features. It is important to note that keystroke features show strong performance when integrated with text and app usage features while also increasing privacy, thereby pushing the Pareto front outwards. It is also interesting to observe that for most models, performance stays level while privacy improves, which is a promising sign for the real-world deployment of such models which requires a balance between both desiderata.

\vspace{-1mm}
\subsection{Qualitative Analysis}
\label{qualitative}
\vspace{-1mm}

To further shed light on the relationships between mood prediction performance and privacy, we performed a more in-depth study of the text, keystroke, and app usage features learned by the model (see Appendix~\ref{qualitative_analysis} for more examples).

\vspace{-1mm}
\textbf{Understanding the unimodal features:} We first analyze how individual words, keystroke timings, and app usage are indicative of positive or negative mood for different users.

\vspace{-1mm}
\textit{Text:} We find that several words are particularly indicative of mood: \textit{can't/cant}, \textit{don't/don't}, and \textit{sorry} are negative for more users than positive, while \textit{yes} is overwhelmingly positive across users (9 pos, 1 neg), but \textit{yeah} is slightly negative (5 pos, 7 neg). We also analyze the use of emojis in typed text and find that while there are certain emojis that lean positive (e.g., \includegraphics[height=1em]{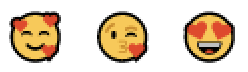}), there are ones (e.g., :( and \includegraphics[height=1em]{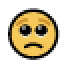}) that used in both contexts depending on the user (see Table~\ref{emojis}).

\definecolor{gg}{RGB}{15,125,15}
\definecolor{rr}{RGB}{190,45,45}

\begin{table}[t!]
\fontsize{8.5}{11}\selectfont
\centering
\caption{Top emojis associated with positive and negative mood (each row is a different user).\vspace{-2mm}}
\setlength\tabcolsep{0.0pt}
\begin{tabular}{c || c}
\Xhline{3\arrayrulewidth}
Positive emojis & Negative emojis \\
\Xhline{3\arrayrulewidth}
\begin{minipage}{.18\textwidth}
    \includegraphics[width=\linewidth]{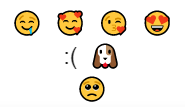}
\end{minipage}
&
\begin{minipage}{.22\textwidth}
    \includegraphics[width=\linewidth]{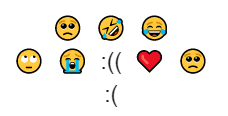}
\end{minipage}\\
\Xhline{3\arrayrulewidth}
\end{tabular}
\label{emojis}
\vspace{-2mm}
\end{table}

\definecolor{gg}{RGB}{15,125,15}
\definecolor{rr}{RGB}{190,45,45}

\begin{table}[t!]
\fontsize{7.5}{11}\selectfont
\centering
\caption{Top $3$ apps associated with positive and negative moods (each row is a different user).\vspace{-2mm}}
\setlength\tabcolsep{2.0pt}
\begin{tabular}{c || c}
\Xhline{3\arrayrulewidth}
Top 3 positive apps & Top 3 negative apps \\
\Xhline{3\arrayrulewidth}
Photos, Settings, Snapchat          & Calendar, Wattpad, SoundCloud \\
FaceTime, MyFitnessPal, Musically   & Notes, App Store, Siri \\
Weather, Phone, FaceTime            & Chrome, App Store, SMS \\
Weather, Phone, Spotify             & Safari, Notes, GroupMe \\
Spotlight, App Store, Uber          & Pinterest, Phone, Yolo \\
Uber, Netflix, LinkedIn             & Phone, Calendar, Safari \\
\Xhline{3\arrayrulewidth}
\end{tabular}
\label{apps}
\vspace{-4mm}
\end{table}

\vspace{-1mm}
\textit{Apps:} In Table~\ref{apps}, we show the top $3$ apps associated with positive or negative moods across several users. It is interesting to observe that many outdoor apps (i.e., \textit{Weather, MyFitnessPal, Uber}), photo sharing apps (i.e., \textit{Photos, Snapchat}), and calling apps (i.e., \textit{FaceTime, Phone}) are associated with positive mood, while personal apps such as personal management (i.e., \textit{Calendar, Notes, Siri}), web browsing (i.e., \textit{Chrome, Safari}), and shopping (i.e., \textit{App Store}) are associated with negative mood. However, some of these findings are rather user-specific (e.g., \textit{Phone} can be both positive or negative depending on the user).

\begin{figure}%
    \centering
    \includegraphics[width=0.7\linewidth]{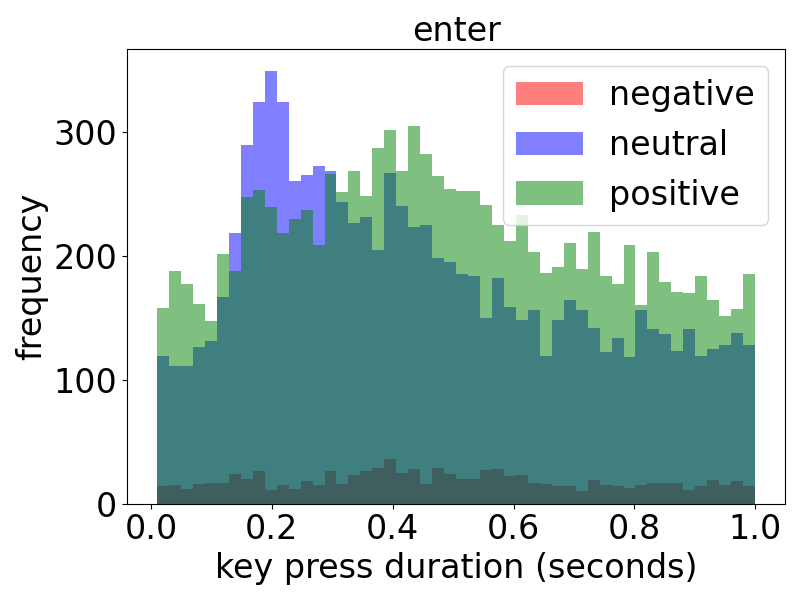}
    \vspace{-2mm}
    \caption{An example where the \textit{`enter'} character keypress is indicative of either positive, neutral, or negative mood depending on the keypress duration.\vspace{-2mm}}
    \label{enter_keystrokes}
\end{figure}

\vspace{-1mm}
\textbf{Understanding the multimodal features:} We also analyze how the \textit{same characters and words} can contribute to \textit{different mood predictions} based on their keystroke patterns. As an example, the distribution of keystrokes for the \textit{enter} character on the keyboard differs according to the daily mood of one user (see Figure~\ref{enter_keystrokes} and Appendix~\ref{qualitative_analysis} for more users).
In Table~\ref{word_keystrokes}, we extend this analysis to entire words. For each of the $500$ most common words, we aggregated their accompanying keystroke timings for user-reported positive and negative mood. These two distributions tell us how the same word in different keystroke contexts can indicate different moods. We performed Wilcoxon rank-sum tests at $5\%$ significance level to compare these distributions and recorded the words in which either faster or slower typing was statistically significantly correlated with either mood. Observe how certain semantically positive words like \textit{love}, \textit{thank}, and \textit{haha} become judged as more positive when typed at a faster speed. Therefore, contextualizing text with their keystroke timings offers additional information when learning representations of typed text.

\definecolor{gg}{RGB}{15,125,15}
\definecolor{rr}{RGB}{190,45,45}

\begin{table}[t!]
\fontsize{8.5}{11}\selectfont
\centering
\caption{Words with significantly different timings associated with positive and negative moods (each row is a different user).\vspace{-2mm}}
\setlength\tabcolsep{2.0pt}
\begin{tabular}{c || c}
\Xhline{3\arrayrulewidth}
Slower implies positive & Faster implies positive \\
\Xhline{3\arrayrulewidth}
just                                    & why, thank, haha \\
next, was, into, people                 & making, work, idk \\
stuff, cute, phone, want, talk, see     & they, send, dont, man, going \\
don't, talk                             & think, you, all, love \\
\Xhline{3\arrayrulewidth}
\end{tabular}
\label{word_keystrokes}
\vspace{-4mm}
\end{table}

\vspace{-1mm}
\section{Conclusion}
\vspace{-1mm}

In this paper, we investigated the learning of language and multimodal representations of \textit{typed text} collected from mobile data. We studied the challenge of learning markers of daily mood as a step towards early detection and intervention of mental health disorders for social good.
Our method also shows promising results in obfuscating user identities for privacy-preserving learning, a direction crucial towards real-world learning from sensitive mobile data and healthcare labels.
In addition, our findings illustrate several challenges and opportunities in representation learning from typed text as an understudied area in NLP.

\vspace{-1mm}
\textbf{Limitations \& future work:} While our approach shows promises in learning representations for mood prediction, several future directions on the modeling and NLP side include: 1) better models and pre-training algorithms for NLP on typed text, 2) algorithms that provide formal guarantees of privacy~\citep{dwork2008differential}, and 3) federated training from decentralized data~\citep{DBLP:journals/corr/McMahanMRA16} to improve privacy~\citep{geyer2017differentially} and fairness~\citep{liang2020fair} of sensitive data. We describe more limitations and future social implications of our work in our broader impact statement in Appendix~\ref{broader_impact}.

\vspace{-1mm}
\section*{Acknowledgements}
\vspace{-1mm}

This material is based upon work partially supported by the National Science Foundation (Awards \#1750439 and \#1734868) and the National Institutes of Health (Award \#U01MH116923). MM is supported by the Swiss National Science Foundation (\#P2GEP2\_184518). RS is supported by NSF IIS1763562 and ONR Grant N000141812861. Any opinions, findings, and conclusions, or recommendations expressed in this material are those of the author(s) and do not necessarily reflect the views of the National Science Foundation, National Institutes of Health, or Office of Naval Research, and no official endorsement should be inferred. We are extremely grateful to Torsten W\"{o}rtwein for helpful discussions and feedback on earlier versions of this paper. Finally, we would also like to acknowledge NVIDIA's GPU support and the anonymous reviewers for their extremely helpful comments.

\bibliography{main}
\bibliographystyle{acl_natbib}

\clearpage
\appendix

\section*{Appendix}

\vspace{-1mm}
\section{Broader Impact Statement}
\label{broader_impact}
\vspace{-1mm}

Learning markers of mood from mobile data presents an opportunity for large-scale adaptive interventions of suicidal ideation. However, there are important concerns regarding its implications to society and policy.

\textbf{Applications in mental health:} Suicide is the second leading cause of death among adolescents. In addition to deaths, $16\%$ of high school students report seriously considering suicide each year, and $8\%$ make one or more suicide attempts~\citep{cdc}. Despite these alarming statistics, there is little consensus concerning imminent risk for suicide~\citep{franklin2017risk,large2017patient}. Current research conducts clinical interviews and patient self-report questionnaires that provide long-term assessments of suicide risk. However, few studies have focused on imminent suicidal risk, which is of critical clinical importance as a step towards adaptive real-time interventions~\citep{glenn2014improving,schuck2019suicide}. Given the impact of suicide on society, there is an urgent need to better understand the behavior markers related to suicidal ideation.

``Just-in-time'' adaptive interventions delivered via mobile health applications provide a platform of exciting developments in low-intensity, high-impact interventions~\citep{nahum2018just}. The ability to intervene precisely during an acute risk for suicide could dramatically reduce the loss of life. To realize this goal, we need accurate and timely methods that predict when interventions are most needed. Monitoring (with participants' permission) mobile data to assess mental health and provide early interventions is, therefore, a rich opportunity for scalable deployment across high-risk populations. Our data collection, experimental study, and computational approaches provide a step towards data-intensive longitudinal monitoring of human behavior. However, one must take care to summarize behaviors from mobile data without identifying the user through personal (e.g., personally identifiable information) or protected attributes (e.g., race, gender). This form of anonymity is critical when implementing these technologies in real-world scenarios. Our goal is to be highly predictive of mood while remaining as privacy-preserving as possible. We outline some of the potential privacy and security concerns below.

\textbf{Limitations:} While we hope that our research can provide a starting point on the potential of detecting mood unobtrusively throughout the day in a privacy-preserving way, we strongly acknowledge there remain methodological issues where \textit{a lot} more research needs to be done to enable the real-world deployment of such technologies. We emphasize that healthcare providers and mobile app startups \textbf{should not} attempt to apply our approach in the real world until the following issues (and many more) can be reliably resolved:
\begin{enumerate}
    \item We do not make broad claims across teenage populations from only $17$ participants in this study. Furthermore, it remains challenging for models to perform person-independent prediction which makes it hard to deploy across large populations.
    \item Our current work on predicting daily mood is still a long way from predicting imminent suicide risk. Furthermore, any form of prediction is still significantly far away from integrating methods like this into the actual practice of mental health, which is a challenging problem involving a broad range of medical, ethical, social, and technological researchers~\cite{resnik2021naturally,lee2021artificial}.
    \item Text and keystrokes can differ for participants who speak multiple languages or non-prestige vernaculars. One will need to ensure that the method works across a broad range of languages to ensure accessibility in its desired outcomes.
    \item This study assumes that participants have no restrictions for data/network connections \& data plans on their phones, which may leave out vulnerable populations that do not meet this criterion.
\end{enumerate}

\textbf{Privacy and security:} There are privacy risks associated with making predictions from mobile data. To deploy these algorithms across at-risk populations, it is important to keep data private on each device without sending it to other locations. Even if data is kept private, it is possible to decode data from gradients~\citep{zhu2020deep} or pretrained models~\citep{carlini2020extracting}. In addition, sensitive databases with private mobile data could be at-risk to external security attacks from adversaries~\citep{lyu2020threats}. Therefore, it is crucial to obtain user consent before collecting device data. In our experiments with real-world mobile data, all participants have given consent for their mobile device data to be collected and shared with us for research purposes. All data was anonymized and stripped of all personal (e.g., personally identifiable information) and protected attributes (e.g., race, gender).

\textbf{Social biases:} We acknowledge that there is a risk of exposure bias due to imbalanced datasets, especially when personal mobile data and sensitive health labels (e.g., daily mood, suicidal thoughts and behaviors, suicide risk). Models trained on biased data have been shown to amplify the underlying social biases especially when they correlate with the prediction targets~\citep{DBLP:journals/corr/abs-1809-07842}. This leaves room for future work in exploring methods tailored for specific scenarios such as mitigating social biases in words~\citep{bolukbasi2016man}, sentences~\citep{liang2020fair}, and images~\citep{10.1145/3209978.3210094}. Future research should also focus on quantifying the trade-offs between fairness and performance~\citep{DBLP:journals/corr/abs-1906-08386}.

Overall, we believe that our proposed approach can help quantify the tradeoffs between performance and privacy. We hope that this brings about future opportunities for large-scale real-time analytics in healthcare applications.

\vspace{-1mm}
\section{Dataset Details}
\label{dataset_details}
\vspace{-1mm}

The Mobile Assessment for the Prediction of Suicide (MAPS) dataset was designed to elucidate real-time indicators of suicide risk in adolescents ages $13-18$ years. Current adolescent suicide ideators and recent suicide attempters along with aged-matched psychiatric controls with no lifetime suicidal thoughts and behaviors completed baseline clinical assessments (i.e., lifetime mental disorders, current psychiatric symptoms). Following the baseline clinical characterization, a smartphone app, the Effortless Assessment of Risk States (EARS), was installed onto adolescents’ phones, and passive sensor data were acquired for $6$-months. Notably, during EARS installation, a keyboard logger is configured on adolescents’ phones, which then tracks all words typed into the phone as well as the apps used during this period. Each day during the $6$-month follow-up, participants also were asked to rate their mood on the previous day on a scale ranging from $1-100$, with higher scores indicating a better mood. After extracting multimodal features and discretizing the labels (see Section~\ref{dataset}), we summarize the final dataset feature and label statistics in Table~\ref{statistics}.

\begin{table*}[h]
\fontsize{9}{11}\selectfont
\centering
\caption{Mobile Assessment for the Prediction of Suicide (MAPS) dataset summary statistics.}
\setlength\tabcolsep{6.0pt}
\begin{tabular}{l | c | c c c | c }
\Xhline{3\arrayrulewidth}
Users & Datapoints & Modalities & Features & Dimensions & Labels \\
\Xhline{3\arrayrulewidth}
\multirow{3}{*}{$17$} & \multirow{3}{*}{$1641$} & Text & bag-of-words, one-hot & $2000$ & \multirow{3}{*}{Daily mood: negative, neutral, positive}\\
& & Keystrokes & bag-of-timings & $100$\\
& & App usage & bag-of-apps, one-hot & $274$ \\
\Xhline{3\arrayrulewidth}
\end{tabular}
\vspace{-0mm}
\label{statistics}
\end{table*}

\vspace{-1mm}
\section{Experimental Setup}
\label{exp_details}
\vspace{-1mm}

We provide additional details on the model implementation and experimental setup.

\vspace{-1mm}
\subsection{Implementation Details}
\vspace{-1mm}

All models and analyses were done in Python. SVM models were implemented with Scikit-learn and MLP/NI-MLP models were implemented with PyTorch. BERT, XLNet, and Longformer models were fine-tuned using Hugging Face (website: https://huggingface.co, GitHub: https://github.com/huggingface).

\vspace{-1mm}
\subsection{Hyperparameters}
\vspace{-1mm}

We performed a small hyperparameter search over the ranges in Table~\ref{params}. This resulted in a total of $35$ hyperparameter configurations for SVM and $12$ for MLP (6 for apps only). By choosing the best-performing model on the validation set, we selected the resulting hyperparameters as shown in Table~\ref{params}.

\begin{table*}[t]
\fontsize{9}{11}\selectfont
\centering
\caption{Model parameter configurations. *Integer kernel values denote the degree of a polynomial kernel.}
\setlength\tabcolsep{6.0pt}
\begin{tabular}{l | c | c }
\Xhline{3\arrayrulewidth}
Model & Parameter & Value \\
\Xhline{3\arrayrulewidth}
\multirow{2}{*}{SVM}
& C & 0.1, 0.5, 1, 2, 3, 5, 10 \\
& Kernel* & RBF, 2, 3, 5, 10 \\
\Xhline{0.5\arrayrulewidth}
\multirow{6}{*}{MLP}
& hidden dim 1 (multimodal \& text only) & 1024, 512 \\
& hidden dim 2 (multimodal \& text only) & 128, 64 \\
& hidden dim 1 (keystrokes only) & 64, 32 \\
& hidden dim 2 (keystrokes only) & 32, 16 \\
& hidden dim 1 (apps only) & 128 \\
& hidden dim 2 (apps only) & 128, 64 \\
& dropout rate & 0, 0.2, 0.5 \\
& learning rate & 0.001 \\
& batch size & 100 \\
& epochs & 200 \\
\Xhline{0.5\arrayrulewidth}
\multirow{2}{*}{\names}
& $\lambda$ & 0.1, 1, 2, 3, 5, 10 \\
& $\sigma$ & 1, 5, 10, 25, 50, 100, 150 \\
\Xhline{3\arrayrulewidth}
\end{tabular}
\label{params}
\end{table*}

\vspace{-1mm}
\subsection{Model Parameters}
\vspace{-1mm}

Each model has about two million parameters. See Table~\ref{params} for exact hidden dimension sizes.

\vspace{-1mm}
\subsection{Training Resources and Time}
\vspace{-1mm}

All experiments were conducted on a GeForce RTX $2080$ Ti GPU with $12$ GB memory. See Table~\ref{times} for approximate running times.

\begin{table*}[!tbp]
\fontsize{9}{11}\selectfont
\centering
\caption{Approximate training times (total across $10$-fold cross validation and hyperparameter search).}
\setlength\tabcolsep{6.0pt}
\begin{tabular}{l | c | c }
\Xhline{3\arrayrulewidth}
Model & Modality & Time (hours) \\
\Xhline{3\arrayrulewidth}
\multirow{6}{*}{SVM}
& Text + Keystrokes + Apps & 10 \\
& Text + Keystrokes & 10 \\
& Text + Apps & 10 \\
& Text & 8 \\
& Keystrokes & 1 \\
& Apps & 1 \\
\Xhline{0.5\arrayrulewidth}
\multirow{6}{*}{MLP (100 epochs, 3 runs)}
& Text + Keystrokes + Apps & 6 \\
& Text + Keystrokes & 5 \\
& Text + Apps & 6 \\
& Text & 5 \\
& Keystrokes & 4 \\
& Apps & 2 \\
\Xhline{0.5\arrayrulewidth}
\multirow{1}{*}{\names}
& all & 4 \\
\Xhline{3\arrayrulewidth}
\end{tabular}
\label{times}
\end{table*}

\vspace{-1mm}
\section{Experimental Details}
\label{extra_results}
\vspace{-1mm}

We present several additional analysis of the data and empirical results:

\vspace{-1mm}
\subsection{Details on Mood Prediction}
\vspace{-1mm}

There is often a tradeoff between privacy and prediction performance. To control this tradeoff, we vary the parameter $\sigma$, which is the amount of noise added to the identity-dependent subspace across batches and training epochs. In practice, we automatically perform model selection using this performance-privacy ratio $R$ computed on the validation set, where
\begin{equation}
    R = \frac{s_{\textsc{MLP}} - s_{\names}}{t_{\textsc{MLP}} - t_{\names}}    
\end{equation}
is defined as the improvement in privacy per unit of performance lost. Here, $s$ is defined as the accuracy in the user prediction task and $t$ is defined as the F1 score on the mood prediction task.

In the rare cases where \names\ performed better than the original \textsc{MLP} and caused $R$ to become negative, we found this improvement in performance always came at the expense of worse privacy as compared to other settings of $\lambda$ and $\sigma$ in \names. Therefore, models with negative $R$ were not considered for Table~\ref{modalities_comparison}.

\vspace{-1mm}
\subsection{Details on Preserving Privacy}
\vspace{-1mm}

\definecolor{gg}{RGB}{15,125,15}
\definecolor{rr}{RGB}{190,45,45}

\begin{table*}[t!]
\fontsize{9}{11}\selectfont
\centering
\caption{Top 5 words associated with positive and negative moods (each row is a different user).\vspace{-0mm}}
\setlength\tabcolsep{6.0pt}
\begin{tabular}{c || c}
\Xhline{3\arrayrulewidth}
Top 5 positive words & Top 5 negative words \\
\Xhline{3\arrayrulewidth}
hot, goodnight, ft, give, keep      & soon, first, ya, friend, leave \\
still, y'all, guys, new, come       & amazing, see, said, idk, look \\
mind, days, went, tf, next          & tired, hair, stg, snap, anyone \\
girls, music, happy, mean, getting  & omg, people, talking, ask, might \\
\Xhline{3\arrayrulewidth}
\end{tabular}
\label{words}
\vspace{-2mm}
\end{table*}

For Table~\ref{privacy_features}, the model with the best privacy out of those within $5\%$ performance of the original MLP model (or, if no such model existed, the model with the best performance) was selected.

Interestingly, in Figure~\ref{fig:tradeoff}, we find that the tradeoff curve on a model trained only using app features does not exhibit a Pareto tradeoff curve as expected. We attribute this to randomness in predicting both mood and identities. Furthermore,~\citet{wang2017select} found that adding noise to the identity subspace can sometimes improve generalization by reducing reliance on identity-dependent confounding features, which could also explain occasional increased performance at larger $\sigma$ values.

Note that we do not include privacy results for features learned by \textsc{SVM}, which finds a linear separator in a specified kernel space rather than learning a representation for each sample. Explicitly projecting our features is computationally infeasible due to the high dimensionality of our chosen kernel spaces.

\definecolor{gg}{RGB}{15,125,15}
\definecolor{rr}{RGB}{190,45,45}

\begin{table*}[t!]
\fontsize{9}{11}\selectfont
\centering
\caption{Top words associated with positive and negative moods across users. We find that while certain positive words are almost always indicative of mood, others are more idiosyncratic and depend on the user.\vspace{-0mm}}
\setlength\tabcolsep{6.0pt}
\begin{tabular}{c c c || c c c}
\Xhline{3\arrayrulewidth}
Positive words & Positive users & Negative users & Negative words & Negative users & Positive users \\
\Xhline{3\arrayrulewidth}
make & 9 & 1 & i'm/im & 10 & 5 \\
yes & 9 & 1 & feel & 7 & 3 \\
got & 7 & 1 & yeah & 7 & 5 \\
still & 7 & 1 & can't/cant & 6 & 2\\
wanna & 7 & 1 & people & 6 & 4 \\
like & 7 & 2 & know & 6 & 4 \\
need & 7 & 2 & go & 6 & 5 \\
send & 7 & 2 & one & 6 & 6 \\
get & 7 & 2 & today & 5 & 1 \\
good & 7 & 3 & day & 5 & 2 \\
\Xhline{3\arrayrulewidth}
\end{tabular}
\label{word_counts}
\vspace{-2mm}
\end{table*}

\begin{figure*}%
    \centering
    \vspace{-0mm}
    \includegraphics[width=\linewidth]{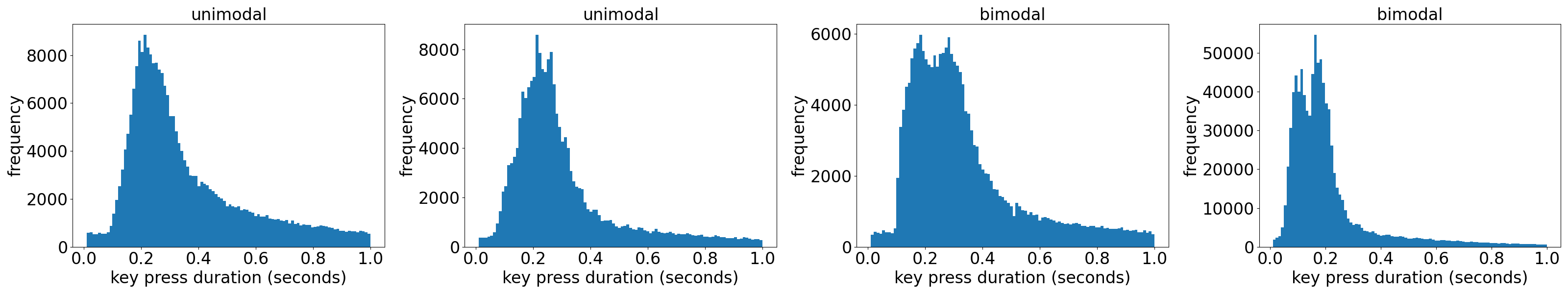}
    \vspace{-2mm}
    \caption{Examples of keystroke timing histograms for different users. We find that the distribution of keystroke timings varies between unimodal and bimodal for different users.\vspace{-2mm}}
    \label{overall_charhist}
\end{figure*}

\vspace{-1mm}
\subsection{Qualitative Analysis}
\label{qualitative_analysis}
\vspace{-1mm}

In this section, we provide more empirical analysis on the unimodal and multimodal features in the MAPS dataset.

\subsubsection{Understanding the unimodal features}

\textit{Text:} We begin with some basic statistics regarding word distributions. For each user, we tallied the frequencies of each word under each daily mood category (positive, neutral, and negative), as well as the overall number of words in each mood category. We define ``positive'' words and emojis to be those with a higher relative frequency of positive mood compared to the overall positive mood frequency, and lower than overall negative mood frequency. Likewise, ``negative'' words and emojis have higher than overall negative mood frequency and lower than overall positive mood frequency. We filtered out words for specific users if the word was used less than $40$ times. Finally, we ranked the words by the difference in relative frequency (i.e., a word is ``more positive'' the larger the difference between its positive mood relative frequency and the user's overall positive mood relative frequency). See Table~\ref{words} for examples of top positive and negative words. For each word, we also counted the number of users for which the word was positive or negative. See Table~\ref{word_counts} for the words with the highest user counts.

\textit{Keystrokes:} We show some sample bag-of-timing histograms in Figure~\ref{overall_charhist}. It is interesting to find that certain users show a bimodal distribution across their keystroke histograms with one peak representing faster typing and another representing slower typing. Visually, the overall keystroke histograms did not differ that much across users which might explain its lower accuracies in both mood and user prediction when trained with \names\ (see Figure~\ref{fig:tradeoff}).

\textit{App usage:} Similar to ``positive'' words, we define ``positive'' apps to be those with higher than overall positive mood relative frequency and lower than overall negative mood relative frequency, and ``negative'' apps to be the opposite. Apps were also then sorted by difference in relative frequency.

\subsubsection{Understanding the multimodal features}

\begin{figure*}%
    \centering
    \includegraphics[width=\linewidth]{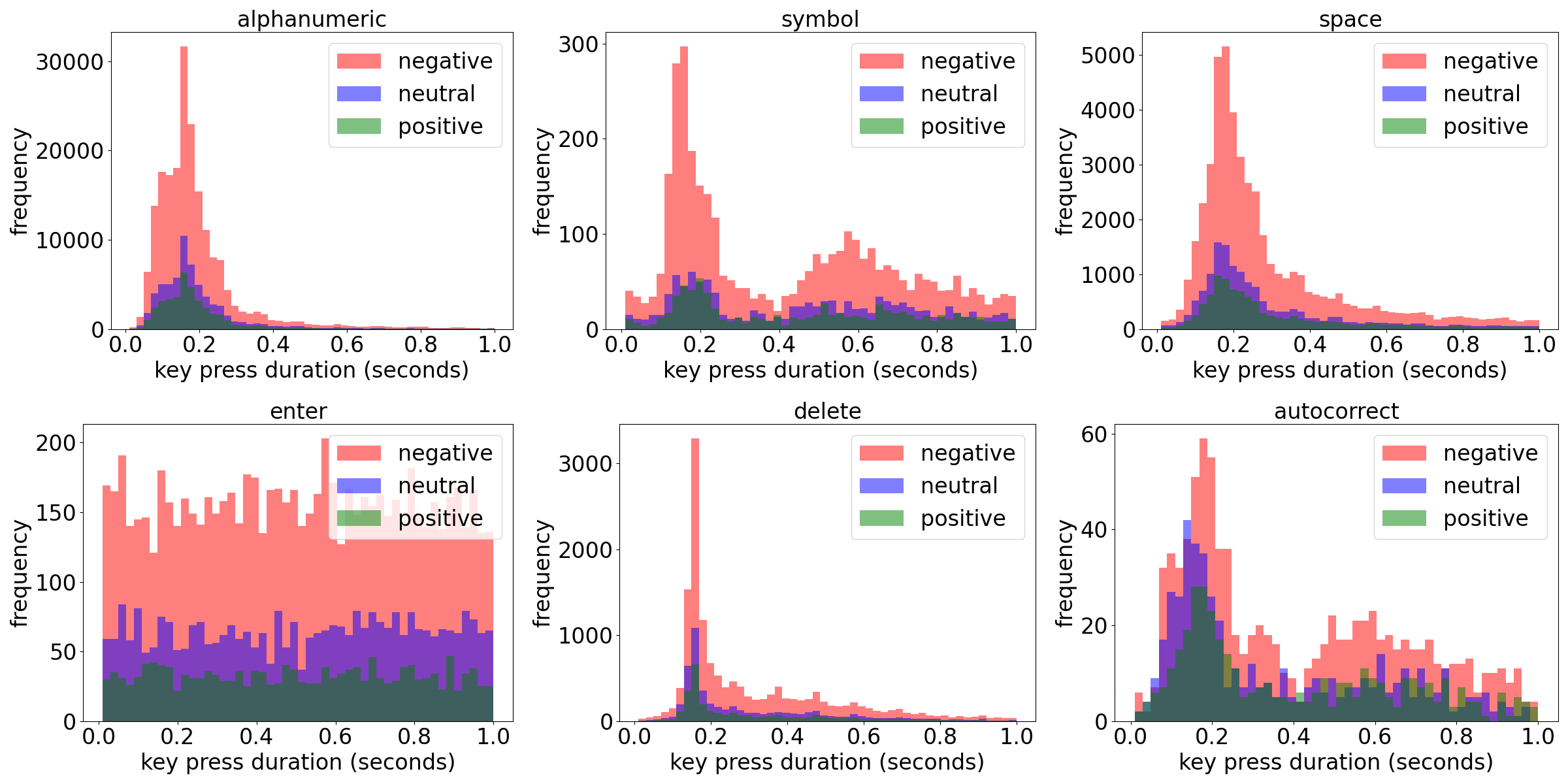}
    \vspace{-2mm}
    \caption{Example of more character key-presses and how their keystroke patterns can be indicative of either positive, neutral, or negative mood. We find particularly interesting patterns in the autocorrect keys and symbols where keystrokes are quite indicative of mood.\vspace{-2mm}}
    \label{more_analysis}
\end{figure*}

\textit{Characters with keystrokes}: For each user, we plotted histograms of keystroke timings of alphanumeric characters, symbols (punctuation and emojis), spacebar, enter, delete, and use of autocorrect, split across daily mood categories. See Figure~\ref{more_analysis} for examples across one user. We find particularly interesting patterns in the autocorrect keys and symbols where keystrokes are quite indicative of mood, which attests to the unique nature of typed text.

\textit{Words with keystrokes}: For each user, we plotted histograms of the word-level keystroke timings of the top $500$ words, split across the daily mood categories of positive, neutral, and negative. We also performed Wilcoxon rank-sum tests at $5\%$ significance level~\cite{wilcoxon1992individual} between the timings of positive and negative mood for each user/word combination to determine which words had significantly different timings between positive and negative mood.

\vspace{-1mm}
\section{Negative Results and Future Directions}
\vspace{-1mm}

Since this is a new dataset, we explored several more methods throughout the research process. In this section we describe some of the approaches that yielded initial negative results despite them working well for standard datasets:

1. \textbf{User specific models:} We also explored the setting of training a separate model per user but we found that there was too little data per user to train a good model. As part of future work, we believe that if \names\ can learn a user-independent classifier, these representations can then be used for further finetuning or few-shot learning on each specific user. Previous work in federated learning~\citep{smith2017federated,liang2020think} offers ways of learning a user-specific model that leverages other users' data during training, which could help to alleviate the lack of data per user.

2. \textbf{User-independent data splits:} We have shown that text, keystrokes, and app usage features are highly dependent on participant identities. Consequently, models trained on these features would perform poorly when evaluated on a user not found in the training set. We would like to evaluate if better learning of user-independent features can improve generalization to new users (e.g., split the data such that the first $10$ users are used for training, next $3$ for validation, and final $4$ for testing). Our initial results for these were negative, but we believe that combining better privacy-preserving methods that learn user-independent features could help in this regard.

3. \textbf{Fine-grained multimodal fusion:} Our approach of combining modalities was only at the input level (i.e., early fusion~\citep{baltruvsaitis2018multimodal}) which can be improved upon by leveraging recent work in more fine-grained fusion~\citep{liang2018multimodal}. One such example could be to align each keystroke feature and app data to the exact text that was entered in, which provides more fine-grained contextualization of text in keystroke and app usage context.

\end{document}